\documentclass[runningheads]{llncs}
\usepackage{graphicx}
\usepackage{url}
\usepackage{tikz}
\usepackage{comment}
\usepackage{amsmath,amssymb} % define this before the line numbering.
\usepackage{color}
\usepackage{booktabs}
\usepackage{enumitem}
\usepackage{mathabx}
\usepackage{wrapfig}
\usepackage[pagebackref,breaklinks,colorlinks]{hyperref}
\usepackage{lipsum}

\def\eg{\emph{e.g.}} 
\def\ie{\emph{i.e.}} \def\Ie{\emph{I.e.}}

\def\thename{PolarDETR}
\def\IMP{Image-based Parametrization}
\def\CAP{Cartesian Parametrization}
\def\POP{Polar Parametrization}
\def\IMPs{Image-based Param.}
\def\CAPs{Cartesian Param.}
\def\POPs{Polar Param.}
\usepackage[accsupp]{axessibility}  % Improves PDF readability for those with disabilities.

\makeatletter\renewcommand\paragraph{\@startsection{paragraph}{4}{\z@}
  {.5em \@plus1ex \@minus.2ex}{-.5em}{\normalfont\normalsize\bfseries}}\makeatother

\begin{document}
% \linenumbers
\pagestyle{headings}
\mainmatter
\def\ECCVSubNumber{-}  % Insert your submission number here

\title{Polar Parametrization for Vision-based \\Surround-View 3D Detection} % Replace with your title

\begin{comment}
\titlerunning{ECCV-22 submission ID \ECCVSubNumber} 
\authorrunning{ECCV-22 submission ID \ECCVSubNumber} 
\author{Anonymous ECCV submission}
\institute{Paper ID \ECCVSubNumber}
\end{comment}
%******************

% \begin{comment}
\titlerunning{Polar Parametrization for Vision-based \\Surround-View 3D Detection}
\author{Shaoyu Chen\inst{1,2} \and
Xinggang Wang\inst{1}\thanks{Xinggang Wang is the corresponding author.}  \and
Tianheng Cheng\inst{1,2}  \and  \\
{Qian Zhang} \textsuperscript{2} \and
{Chang Huang} \textsuperscript{2} \and
{Wenyu Liu} \textsuperscript{1}}

\authorrunning{Chen et al.}
\institute{Huazhong University of Science \& Technology \and
 Horizon Robotics  \\
\email{\{shaoyuchen,xgwang,thch,liuwy\}@hust.edu.cn}\\
\email{\{qian01.zhang, chang.huang\}@horizon.ai}}
% \end{comment}
%******************
\maketitle

\begin{abstract}
3D detection based on surround-view camera system is a critical technique in autopilot. 
In this work, we present \POP\ for 3D detection, which 
reformulates position parametrization, velocity decomposition, perception range, 
label assignment and loss function in polar coordinate system.
\POP\ establishes explicit associations between image patterns and
prediction targets, exploiting
the view symmetry of surround-view cameras as inductive bias 
to ease optimization and boost performance. Based on \POP, 
we propose a surround-view 3D DEtection TRansformer, 
named \thename. \thename\ achieves promising performance-speed trade-off on different backbone configurations.
Besides, \thename\  ranks 1st on the leaderboard of nuScenes benchmark 
in terms of both 3D detection and 3D tracking at the submission time (Mar. 4th,\ 2022). Code will be released at \url{https://github.com/hustvl/PolarDETR}.
\end{abstract}

\section{Introduction}

In the field of autopilot, surround-view camera system has been attached great importance and popularized by both industry and academia, given its low assembly cost and rich semantic information.
And 3D detection based on such vision system has become
the critical technique of environmental perception.
The passed several years have witnessed tremendous  progress in  vision-based 3D detection~\cite{Mono1,Mono2,Mono3,FCOS3D,DD3D,PGD,DETR3D,BEVDet}. Most methods parameterize object's position in two manners, 1) \IMP\ ( \IMPs\ for short) and 2) \CAP\ (\CAPs\ for short).

\begin{figure}[]
    \centering
    \includegraphics[width=\linewidth]{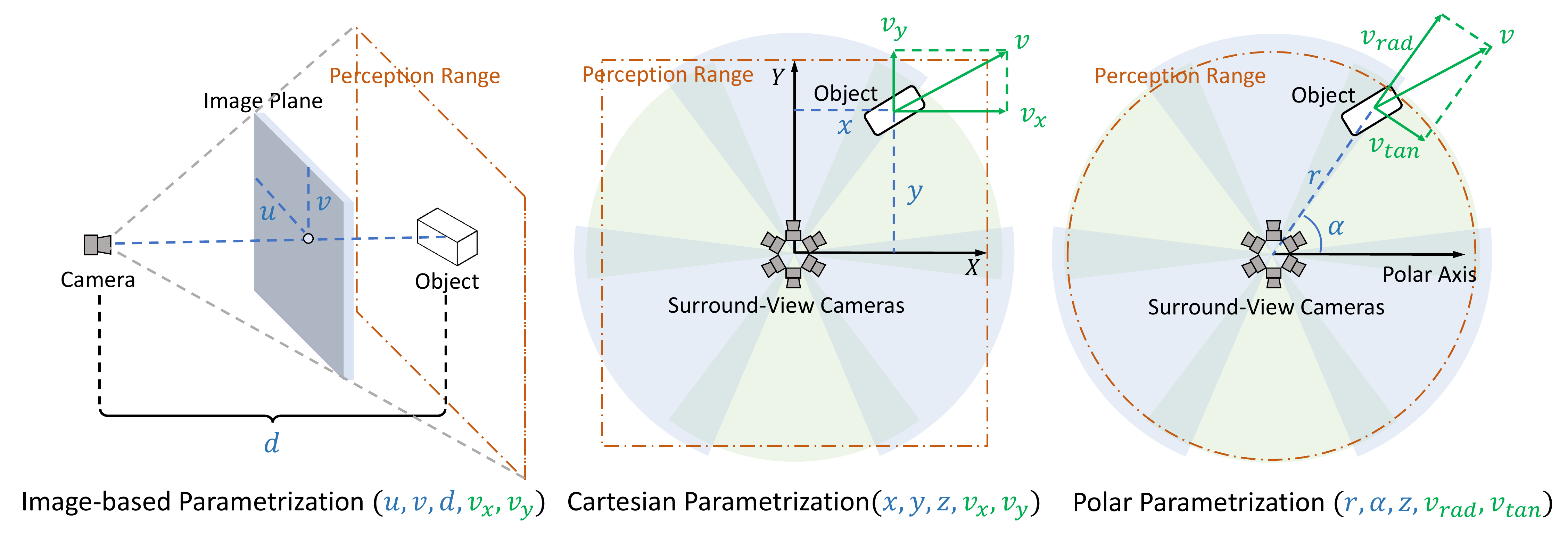}
    \caption{ \textbf{Illustration of parametrization of objects.}
    Object's position is parameterized by pixel coordinate with depth $(u, v, d)$ in \IMP, 
    by cartesian coordinate $(x, y, z)$ in \CAP,
    and by polar (cylindrical) coordinate $(r, \alpha, z)$ in \POP.
    Besides, \POP\ decomposes velocity along radial and tangential direction (to $v_{rad}$ and $v_{tan}$), and reformulates perception range, label assignment and loss function in polar coordinate system.
    Sector denotes field of view (FoV) of each camera.}
    \label{fig:Param}
\end{figure}

\paragraph{\textbf{\IMP.}}
For \IMPs\ (Fig.~\ref{fig:Param} left), object's position is defined in the pixel coordinate frame, parameterized by a $3$-tuple $(u, v, d)$, where $(u, v)$ is the pixel coordinate and $d$ is the object's depth relative to camera. 
With camera's extrinsic and intrinsic parameters, we can transform
$(u, v, d)$ to a 3D point to localize the object in 3D space.
\IMPs\ is widely used in monocular methods~\cite{FCOS3D,DD3D,PGD,SMOKE}. To process surround-view images, they independently take each view's image as input to regress object's $(u, v, d)$. Predicted objects of different views are then projected to the same 3D space for inter-camera merging. Post-processing among views (\eg, NMS) is adopted to filter out duplicated predictions.
However, \IMPs\ raises some problems:
\begin{itemize}[itemsep=2pt,topsep=2pt,parsep=0pt]
\item
Estimating depth from single image is inherently an ill-posed inverse problem. 
The predicted depth is of large error and thus object's positioning accuracy is poor.
For the surround-view camera system, adjacent views overlap with each other. The correlation among views can be leveraged to promote perception performance, which is neglected in \IMPs
\item
Post-processing among views is tricky and unstable.  When the predictions from different views  do not overlap in 3D space, NMS fails to filter out duplicated predictions.
\end{itemize}

\paragraph{\textbf{\CAP.}}
For \CAPs\ (Fig.~\ref{fig:Param} mid), object's position is parameterized by 3D cartesian coordinate $(x, y, z)$. Correspondingly, the perception range is a rectangular region, denoted as $\{(x, y), |x| < X_{\max}, |y| < Y_{\max} \}$.
\CAPs\ is adopted in some recent studies~\cite{BEVDet,DETR3D}. 
Compared with monocular methods, they consider the correlation among views, \ie, taking all views together as input to jointly regress object's 3D coordinate $(x, y, z)$. 

\begin{wrapfigure}[18]{R}{0.45\linewidth}
% \begin{figure}[]
    \centering
    \vspace{-23pt}
    \includegraphics[width=\linewidth]{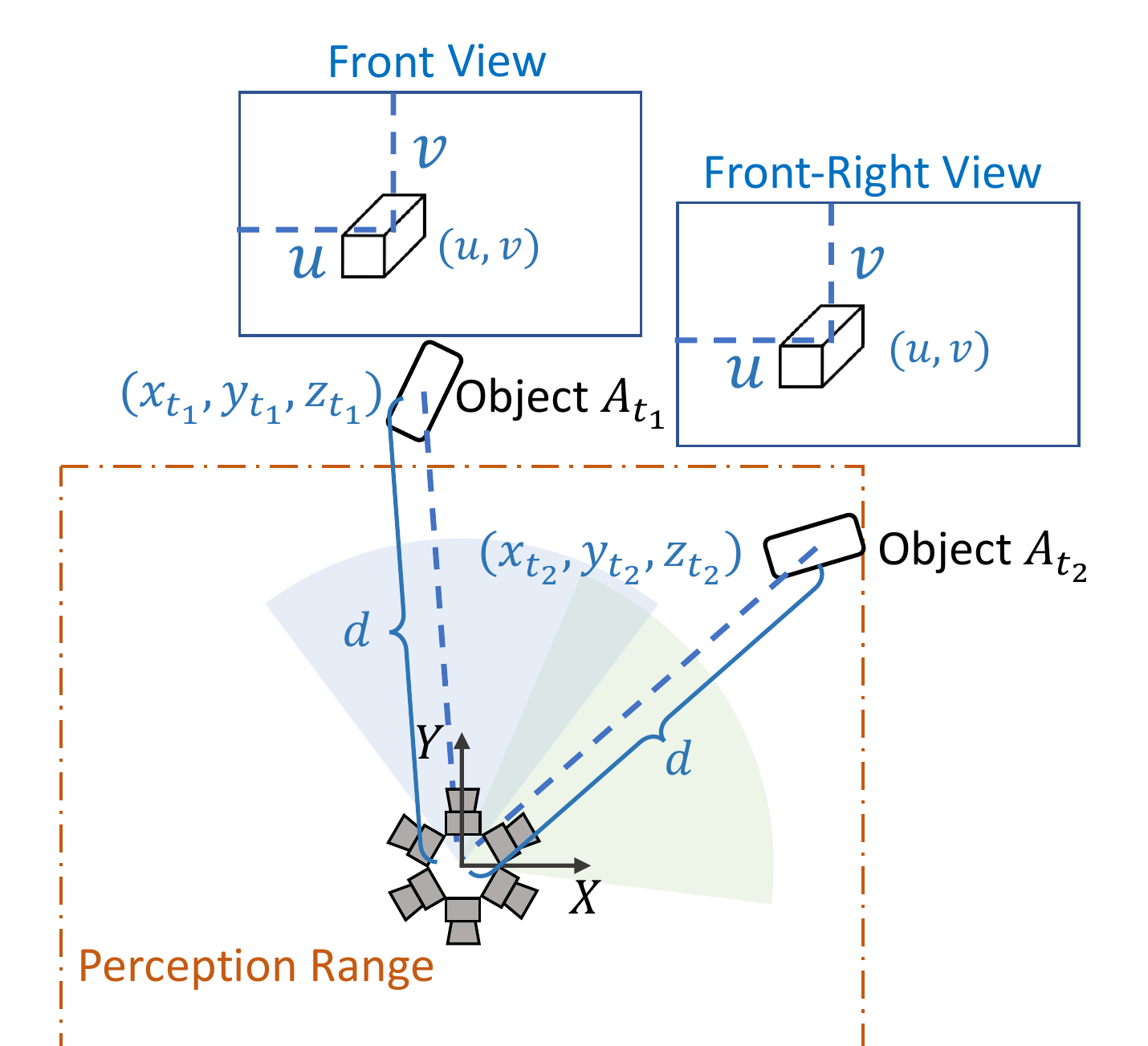}
    \caption{\textbf{A special case for illustrating the problems of \CAPs} Assume  $A_{t_1}$ and $A_{t_2}$ are totally the same in image, except appearing in different views.}
    \label{fig:Problem}
% \end{figure}
\end{wrapfigure}

But \CAPs\ also raise problems. 
We take a special case for example to illustrate them. As shown in Fig.~\ref{fig:Problem}, 
assume that object $A$ appears in different views at timestamp $t_1$ and $t_2$. They have the same pixel coordinate $(u, v)$ and  radial distance $d$.
Their image patterns are also the same.
\begin{itemize}[itemsep=2pt,topsep=2pt,parsep=0pt]
\item
\CAPs\ leads to ambiguity in label assignment because of its rectangular perception range. 
Ground-truth objects out of the perception range are ignored in label assignment. In the case of Fig.~\ref{fig:Problem}, though $A_{t_1}$ and $A_{t_2}$ are at the same distance $d$ and have the same image patterns, $A_{t_1}$ is ignored while $A_{t_2}$ is kept.
$A_{t_1}$ and $A_{t_2}$ are not treated equally in the training phase,
which is contradictory and harmful to the convergence of network. 
\item
\CAPs\ neglects the view symmetry of surround-view cameras.
From the perspective of machine learning, the detector aims at approximating a function $F$ that maps from image patterns $X$ to the prediction targets $Y$, \ie, $Y = F(X)$.
In the case of Fig.~\ref{fig:Problem}, 
for \IMPs,  $A_{t_1}$ and $A_{t_2}$ have the same image patterns $X$  and prediction targets $(u, v, d)$, and can be treated as the same mapping in $F$.
$A_{t_1}$ and $A_{t_2}$ are symmetrical in terms of view.
\IMPs\  leverages the view symmetry as inductive bias of the detector, making it easier to approximate $F$.
While for \CAPs, object $A_{t_1}$ and $A_{t_2}$ correspond to different prediction targets ($(x_{t_1}, y_{t_1}, z_{t_1})$ for $A_{t_1}$ and $(x_{t_2}, y_{t_2}, z_{t_2})$ for $A_{t_2}$).
Since the view symmetry is neglected,
the mappings for $A_{t_1}$ and $A_{t_2}$ are learned separately. Thus, $F$ is more complicated and optimization becomes harder.
\end{itemize}

In this work, we propose a surround-view 3D DEtection TRansformer, named \thename.
To adapt to the view symmetry of surround-view camera system, we parameterize object's position by polar (cylindrical)\footnote{'Cylindrical' is more proper for describing 3D space. But considering  'polar' better reflects the insight of this work and  Z dimension is not critical, we adopt 'polar' for method description} coordinate $(r, \alpha, z)$ and decomposes velocity to radial velocity $v_{rad}$ and tangential velocity $v_{tan}$.
Besides, we reformulate perception range, label assignment and loss function in polar coordinate system.
We term it as \POP\ (\POPs\ for short).

\POPs\ establishes explicit associations between image patterns and prediction targets.
Radial distance $r$ is associated with  object's size in image.
Azimuth $\alpha$ is associated with the index of pixel.
We add 3D positional encoding to each pixel which contains explicit clues about azimuth.
% Thus, azimuth can be directly learned from the encoding.
Radial velocity $v_{rad}$ is associated with the changing rate of object's size.
Tangential velocity $v_{tan}$ is associated with the  object's movement in image plane (similar to optical flow).
With these explicit associations, the mapping is simplified and the  detector achieves better convergence and performance.

Besides, \thename\ enables center-context feature aggregation to enhance the information interaction between object queries and images, and adopts pixel ray as positional encoding to provide 3D spatial priors and help to predict azimuth $\alpha$.

We evaluate \thename\ on the challenging nuScenes~\cite{nuScenes} benchmark.
\thename\ achieves  promising performance-speed trade-off on different backbone configurations. 
And we submit \thename\ for official evaluation on the nuScenes \textit{test} set.
\thename\  \textbf{ranks 1st} on  the highly competitive 3D detection and tracking leaderboard at the submission time (Mar. 4th,\ 2022). 
Besides, thorough ablation studies are provided to validate the effectiveness of \thename.

\section{Related Work}
\subsection{Polar Coordinate}
Polar coordinate is leveraged in some previous works for data representation.
In the field of point cloud semantic segmentation, some grid-based methods~\cite{PolarNet,Cylinder3D}
partition 3D space into polar grids and 
transform LiDAR point clouds into the grid representation, in order to ease the problem of long-tailed distribution.
In the field of image instance segmentation, PolarMask~\cite{PolarMask} formulates the instance mask as a set of contour points represented in the polar coordinate system centered at the instance.
Differently, in order to adapt to the view symmetry of surround-view cameras, \thename\ parameterizes object's position in polar coordinate system and reformulates label assignment and loss function accordingly.

\subsection{DETR-based Object Detection}
Recently, DETR~\cite{DETR} formulates object detection as a set prediction problem and exploits a standard transformer~\cite{TransformerVaswaniSPUJGKP17}, which contains a simple encoder-decoder structure without hand-craft designs and achieves promising performance.
Based on Hungarian algorithm~\cite{HungarianStewartAN16}, the set of object queries can be matched with targets in a one-to-one manner, which helps remove the non-maximum suppression (NMS).
Deformable DETR~\cite{DeformDETR} motivated by \cite{DCNv2} adopts multi-scale deformable attention in the transformer and achieves faster convergence and better performance than DETR.
\cite{CondDETR,SMCA} also address the convergence problem in DETR by incorporating spatial information into object queries.
Several works~\cite{pnpdetr,SparseDETR} propose to prune the redundant tokens in transformers to minimize the computation cost of DETR.
In terms of 3D object detection, DETR3D~\cite{DETR3D} extends DETR  for 3D domain.
Inspired by Deformable DETR, DETR3D projects 3D object queries to 2D reference points to aggregate features from all views.
DETR3D builds up a simple DETR-based pipeline for 3D object detection, while it suffers from (1) ambiguity in label assignment, (2) neglecting view symmetry and (3) insufficient contextual information.
To solve these problems,
\thename\ adopts \POP\ to ease optimization and  enables center-context feature aggregation to enhance the feature interaction.

\subsection{Vision-based 3D Object Detection}
Vision-based 3D detection is a basic perception task in autonomous driving. Early studies are mainly based on KITTI~\cite{KITTI} dataset, which provides  front-view object annotations. KITTI boosts the development of monocular 3D object detection methods~\cite{Mono1,Mono2,Mono3,Mono4,Mono5,Mono6,Mono7,Mono8}.
Recently, with nuScenes~\cite{nuScenes} dataset available,  which contains $360^\circ$ annotations around the ego vehicle, new 3D detection paradigms have been proposed.
Some works~\cite{FCOS3D,DD3D,PGD} still follows the monocular pipeline to detect objects, and  then project the multi-view detection results to the same coordinate system and adopt NMS to merge results.
DETR3D~\cite{DETR3D} extends DETR for 3D object detection.
BEVDet~\cite{BEVDet} projects surround-view image features to Bird-Eye-View (BEV) space, and set detection head on BEV features.
We presents a new paradigm specially designed for surround-view camera systems, in which the view symmetry is exploited to ease optimization and boost performance.

\section{\thename}
\sloppy
\subsection{Overview}
\begin{figure}[t]
    \centering
    \includegraphics[width=\linewidth]{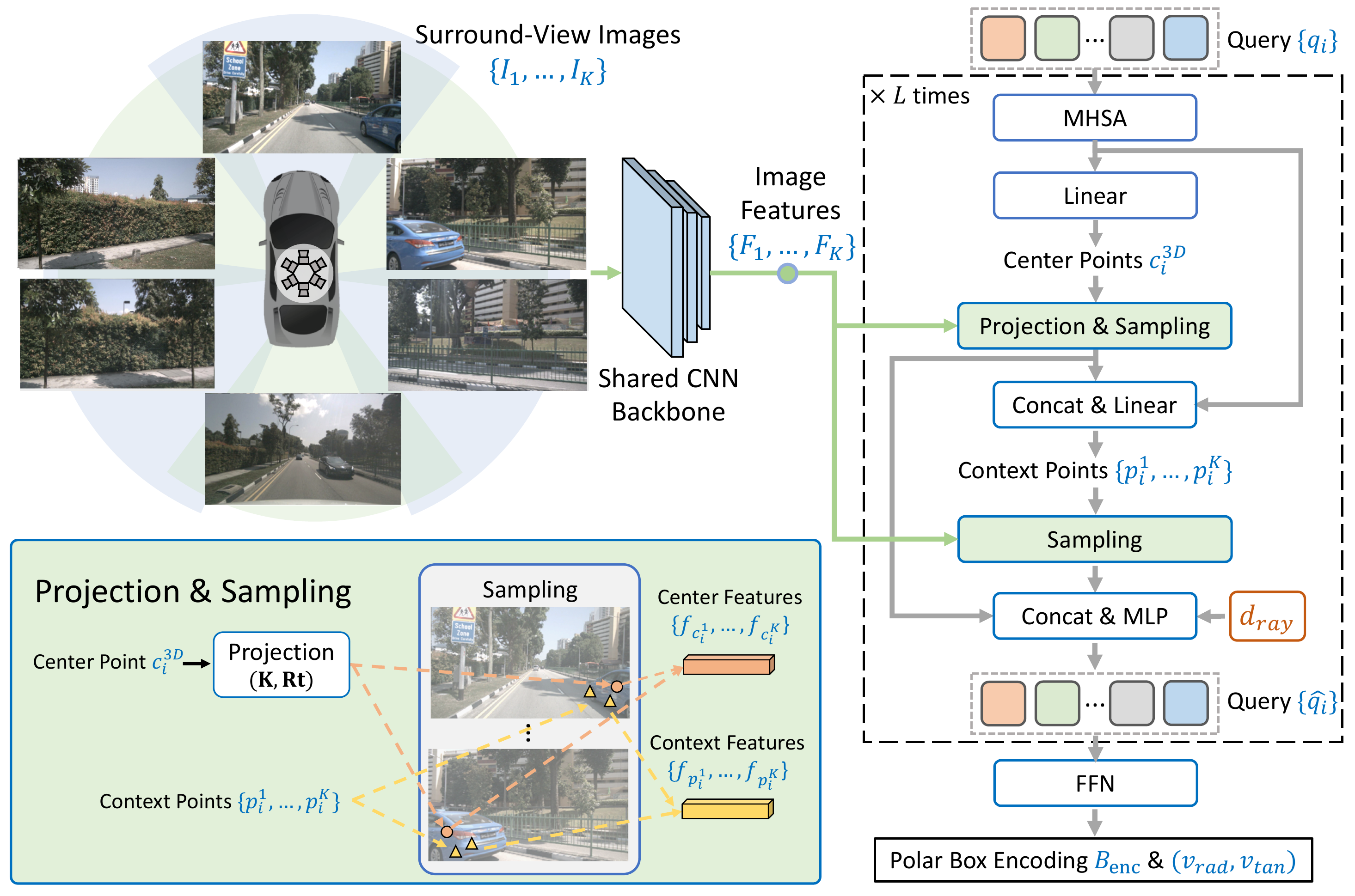}
    \caption{\textbf{The framework of \thename.}
    A shared CNN backbone extracts image features $\mathcal{F}_{\text{img}}=\{{F}_1, \dots, {F}_K\}$ from surround-view images $\mathcal{I}=\{{I}_1, \dots, {I}_K\}$. 
    Decoder layers aggregate center and context features $\{f_{c_i^1},...,f_{c_i^K}, f_{p_i^1},...,f_{p_i^K}\}$ from surround-view feature maps and  update queries $\{q_i\}$ iteratively. Pixel ray  $d_{\text{ray}}$ is introduced to provide spatial priors. FFN predicts polar box encodings $B_{\text{enc}}$, polar velocity components $(v_{rad}, v_{tan})$, and class labels.}
    \label{fig:Framework}
\end{figure}

Fig.~\ref{fig:Framework} illustrates the proposed \thename, which follows the DETR~\cite{DETR,DeformDETR,DETR3D} paradigm.
Given surround-view images $\mathcal{I}=\{{I}_1, \dots, {I}_K\}$ (K denotes the number of views), a shared CNN backbone extracts image features $\mathcal{F}_{\text{img}}=\{{F}_1, \dots, {F}_K\}$. 
A set of object queries $\mathcal{Q}=\{{q}_1, \dots, {q}_N\}$ (N denotes the number of queries) is used to detect objects.
Specifically, each object query encodes the semantic features and positional information of the corresponding object.
And a series of decoder layers aggregate features from surround-view feature maps and  update queries iteratively. 
The feed forward network (FFN) follows the decoder layers and predicts polar box encodings $B_{\text{enc}}$, polar velocity components $(v_{rad}, v_{tan})$, and class labels based on queries.

\subsection{\POP}
In \thename, object's position is parameterized by the polar coordinate.
FFN outputs a $9$-tuple polar box encoding, denoted as,
\begin{equation}
\begin{aligned}
B_{\text{enc}} = (b_r, b_{\sin\alpha}, b_{\cos\alpha}, b_z, b_l, b_w, b_h, b_{\sin\theta}, b_{\cos\theta}). 
\end{aligned}
\end{equation}
We then decode $B_{\text{enc}}$ to get the predicted box vector $B_{\text{pred}}$ represented in polar coordinate system, \ie,
\begin{equation}
\label{eq:transform}
\begin{aligned}
    & \ \ \ \ \ B_{\text{pred}} = (r, \sin\alpha, \cos\alpha, z, l, w, h, \sin\theta, \cos\theta), \\
    & r = \sigma(b_r) \cdot R_{\max},\ \  z = \sigma(b_z) \cdot (Z_{\max} - Z_{\min}) + Z_{\min}, \\
    & \sin\alpha = \frac{b_{\sin\alpha}}{\sqrt{{b_{\sin\alpha}}^2 + {b_{\cos\alpha}}^2}}, \ \ 
    \cos\alpha = \frac{b_{\cos\alpha}}{\sqrt{{b_{\sin\alpha}}^2 + {b_{\cos\alpha}}^2}}, \\
    & \ \ \ \ \ \ l = \exp(b_l), \ \ w = \exp(b_w),\ \  h = \exp(b_h), \\
    & \sin\theta = \frac{b_{\sin\theta}}{\sqrt{{b_{\sin\theta}}^2 + {b_{\cos\theta}}^2}}, \ \ 
    \cos\theta = \frac{b_{\cos\theta}}{\sqrt{{b_{\sin\theta}}^2 + {b_{\cos\theta}}^2}}, \\
\end{aligned}
\end{equation}
where $r$, $\alpha$, $z$ are the radial distance, azimuth and height, indicating the position of object's geometry center in  3D space.
$(l, w, h)$ and $\theta$ denote the size and orientation of bounding box, respectively.
$Z_{\max}$ and $Z_{\min}$ denote the perception range along the $Z$ dimension. $R_{\max}$ denotes the maximum perception distance. $\sigma$ is sigmoid function.
To make sure the continuity of the regression space~\cite{Continuity}, we parameterize both $\alpha$ and $\theta$ by a 2-D $(\sin(\cdot), \cos(\cdot))$ pair.

For \POPs, object's position is decoupled into the radial distance and the azimuth.
The radial distance is symmetrical in different views. It's highly correlated with the object size in image, and can be learned from image patterns.
The azimuth is relative to the indices of cameras and pixels which capture the object, and can be learned from the positional encodings.
Compared with \CAPs, which localizes objects by regressing the cartesian coordinate $(x,y)$, \POPs{} makes more sense.

\paragraph{\textbf{Polar Decomposition for Velocity Estimation.}}
Most methods decompose velocity along the cartesian axes 
and get velocity components $v_x$ and $v_y$.
Differently, \thename\ decomposes 
velocity to radial velocity $v_{rad}$ and tangential velocity $v_{tan}$.
Radial velocity $v_{rad}$  is associated with the changing rate of
object's size. Tangential velocity $v_{tan}$ is associated with
the object's movement in image plane (similar to optical flow).
\thename\ establishes explicit associations between image patterns (input) and
velocity (prediction target), resulting in more accurate velocity estimation.

\subsection{Decoder Layer}
Decoder layers iteratively aggregate features and update queries.
Each decoder layer begins with a multi-head self-attention module (MHSA) for inter-query information interaction.
Then a linear layer extracts object's 3D position from each query,
\ie,
\begin{equation}
\begin{aligned}
     (b_r, b_{\sin\alpha}, b_{\cos\alpha}, b_z) = \text{Linear}(\text{MHSA}(q_i)). \\
\end{aligned}
\end{equation}
We decode $(b_r, b_{\sin\alpha}, b_{\cos\alpha}, b_z)$ with Eq.~(\ref{eq:transform}) and get object's 3D center point $c_i^{\text{3D}}=(r, \alpha, z)$.
\paragraph{\textbf{Center-Context Feature Aggregation.}} 
Then we aggregate features from surround-view feature maps through a two-step center-context procedure.

Firstly, we project the 3D center point $c_i^{\text{3D}}$ to each image and get a set of 2D center points $\{c_i^1, ..., c_i^K\}$, which denote object's centers in all views. \Ie, 
\begin{equation}
	\label{eq:mapping}
    c_i^k = \mathbf{K}^k \cdot \mathbf{Rt}^k \cdot c_i^{\text{3D}},  
\end{equation}
where $\mathbf{K}^k$ and $\mathbf{Rt}^k$ are the projection matrices of view $k$  derived from camera's intrinsics and extrinsics respectively.
Based on $\{c_i^1, ..., c_i^K\}$, we sample center features $\{f_{c_i^1},...,f_{c_i^K}\}$ from surround-view feature maps $\{{F}_1, \dots, {F}_K\}$ through bilinear interpolation. 
A 3D point may be invisible in some views and the projected 2D points are out of range of the images. In this case, the corresponding point features are set to zero.

Considering that center features are not informative enough for localizing objects, we further include context features to enhance the interaction between queries and surround-view features.
Specifically, based on both the center features $\{f_{c_i^1},...,f_{c_i^K}\}$ and the query embeddings $q_i$,
we generate a set of context points $\{p_i^1,..., p_i^K\}$ by predicting the offset relative to the center points.
\Ie,
\begin{equation}
    \Delta u_i^k, \Delta v_i^k  = \text{Linear}(\text{Concat}(f_{c_i^k}, q_i)), \ \ 
    p_i^k = c_i^k + (\Delta u_i^k, \Delta v_i^k).
    \label{eq:offset}
\end{equation}
For clarity, in notation we only consider one context point for each view to avoid introducing extra subscripts. In our implementation, more context points are adopted for better context modeling.
Similarly, based on context points $\{p_i^1,..., p_i^K\}$, we sample context features  $\{f_{p_i^1},...,f_{p_i^K}\}$ from surround-view image features $\{{F}_1, \dots, {F}_K\}$ through bilinear interpolation.

\begin{wrapfigure}[15]{R}{0.45\linewidth}
    \centering
    \includegraphics[width=\linewidth]{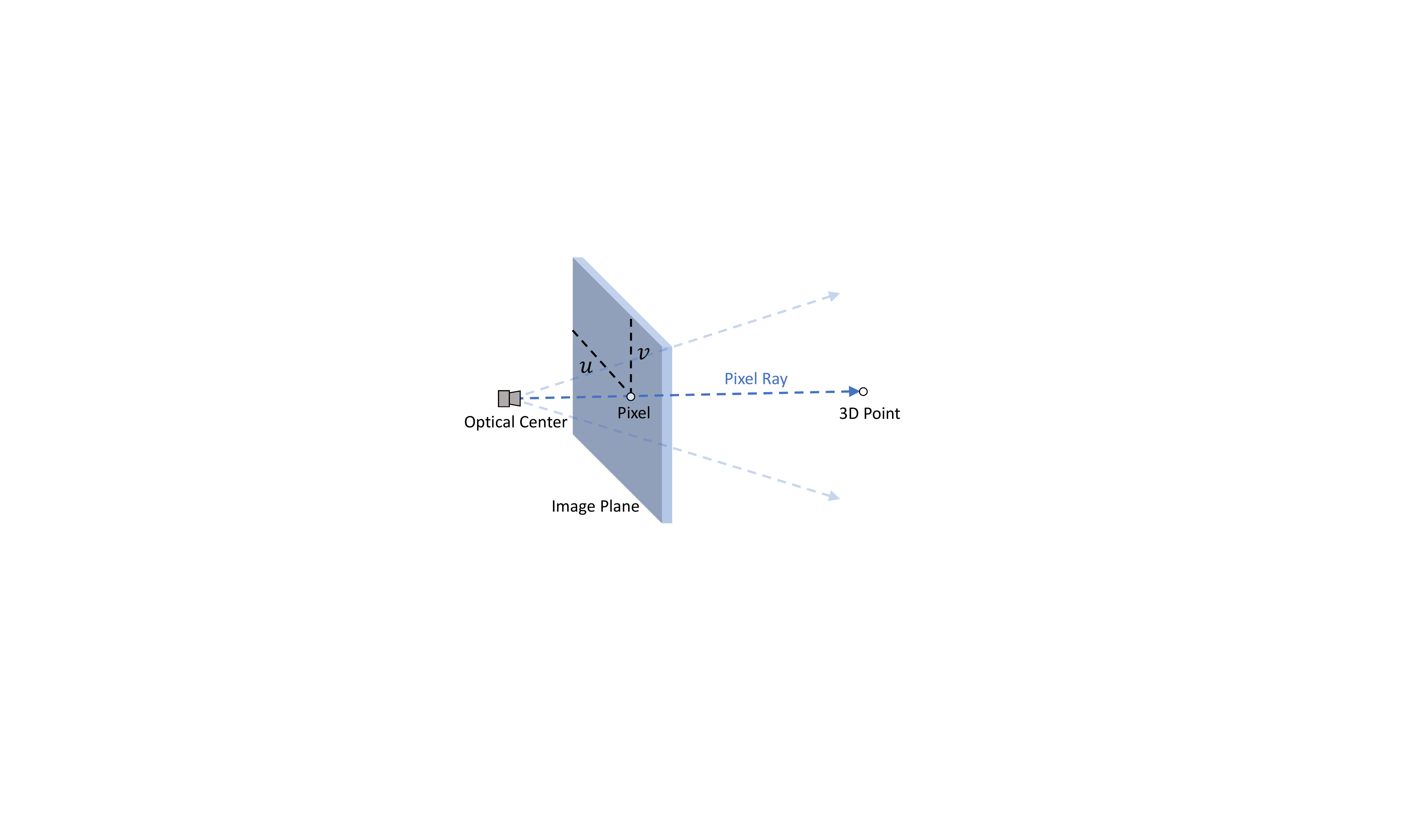}
    \caption{\textbf{Illustration of pixel ray.} Pixel ray travels from camera's optical center through the pixel to the  corresponding 3D point, which encodes the correspondence between 2D image pixel and 3D point.}
    \label{fig:Ray}
\end{wrapfigure}

\paragraph{\textbf{Pixel Ray.}}
Motivated by \cite{NERF,mvp}, we introduce pixel ray as 3D spatial priors. As shown in Fig.~\ref{fig:Ray}, pixel ray travels from camera's optical center through the pixel to the corresponding 3D point.
It encodes the correspondence between 2D image pixel and 3D point, and contains explicit clues about azimuth. 
We leverage pixel ray as extra positional encodings.
Specifically, for each center or context point, the unit direction vector of corresponding pixel ray $d_{\text{ray}}$ is concatenated with point features in the channel dimension (refer to Eq.~(\ref{eq:update})).

\paragraph{\textbf{Query Update.}}
Then we aggregate center and context features with $d_{\text{ray}}$ to update query embeddings, \ie,
\begin{equation}
    \label{eq:update}
    \hat{q_i} = \text{MLP}( \text{Concat}( \{f_{c_i^1},...,f_{c_i^K}, f_{p_i^1},...,f_{p_i^K}\}, d_{\text{ray}})) + q_i.
\end{equation}
The updated query embeddings encode more accurate positional information of the object and contribute to better feature aggregation in the next decoder layer.

\subsection{Perception Range, Label Assignment and Loss Function}
\POPs\ reformulates perception range,  label assignment and loss function in polar coordinate system.
\paragraph{\textbf{Perception Range.}}
As discussed above, for \CAPs\ the rectangular perception range  introduces ambiguity into label assignment.
For \POPs,  the concerned perception range is a circular region with radius $R_{\max}$ centered at the ego vehicle. It avoids ambiguity and fits with the common sense.
It's worth noting that the evaluation region of nuScenes benchmark~\cite{nuScenes} corresponds with the circular perception range of \POPs
\paragraph{\textbf{Label Assignment.}}
Label assignment between predictions and ground-truth (GT) objects is also performed based on polar coordinate.
3D box annotations are first transformed to polar coordinate representation, \ie,
\begin{equation}
    B_{gt} = (\bar{r}, \bar{\sin\alpha}, \bar{\cos\alpha}, \bar{z}, \bar{l}, \bar{w}, \bar{h}, \bar{\sin\theta}, \bar{\cos\theta}).
\end{equation}
Then we adopt bipartite matching to uniquely assign predictions with ground-truth boxes. Assume there exist $N$ predictions and $M$ GTs.
The pair-wise matching cost between prediction $i$ and GT $j$ is defined as,
\begin{equation}
% \centering
\begin{aligned}
     & C(i, j) = C_{\text{cls}}(i, j) + C_{\text{box}}(i, j), \\
     C_{\text{box}}(i, j) =  |r & - \bar{r}| + k_{\text{scaling}} \cdot (|\sin\alpha - \bar{\sin\alpha}| + |\cos\alpha - \bar{\cos\alpha}|), \\
\end{aligned}
\end{equation}
where $C_{\text{cls}}(i, j)$ is the class term inherited from  DETR~\cite{DETR}.
$C_{\text{box}}(i, j)$ is the box term based on \POPs\ 
$k_{\text{scaling}}$ is the scaling factor, which numerically scales up the value of azimuth.
In radial direction,   $r \in [0, R_{\max}]$ ($R_{\max}$ is set to $50$ in nuScenes).  
In tangential direction, $\sin\alpha, \cos\alpha \in [-1,1]$.
Their values differ by more than an order of magnitude. Without $k_{\text{scaling}}$, the radial direction will dominate the assignment and there would be significant assignment error in the tangential direction.
By computing the cost of each prediction-GT pair, we get the cost matrix ${\mathbf{H}} \in \mathbb{R}^{M \times N}$. Then Hungarian algorithm~\cite{HungarianStewartAN16} is adopted to find the optimal assignment.

\paragraph{\textbf{Loss Function.}}
We reformulate the loss function based on polar coordinate.
The bipartite matching loss consists of two parts: a focal loss~\cite{FocalLoss} for class labels and a $\mathcal{L}_1$ loss for polar box parameters $(r, \sin\alpha, \cos\alpha, z, l, w, h, \sin\theta, \cos\theta)$ and polar velocity components $(v_{rad}, v_{tan})$. 
The terms about azimuth, \ie,
$\sin\alpha$ and $\cos\alpha$, are also scaled up by $k_{\text{scaling}}$ to
balance the error distribution between tangential and radial direction.

\subsection{Temporal Information}
Temporal information is important for velocity estimation and occlusion cases.
We straightforwardly and elegantly extend \thename\ to \thename-T, which
takes streaming camera frames as input.
The 3D center point $c_i^{\text{3D}}$ are projected to  past  frames for fetching image features.
Taking frame $t-n$ for example,
\begin{equation}
    c_i^{k\ (t-n)} = \mathbf{K}^k \cdot \mathbf{Rt}^k  \cdot \mathbf{Pose}^{(t-n)} \cdot c_i^{\text{3D}},  
\end{equation}
where $c_i^{k\ (t-n)}$ denotes the corresponding 2D point in the  frame $t-n$,
and $\mathbf{Pose}^{(t-n)}$ denotes the pose transformation matrix which reflects the movement of the ego-vehicle in the time interval $[t-n, t]$.
We sample center and context features from past frames in the same manner of current frame as mentioned above. And all the sampled features (both current and past ones) are aggregated together to update query embeddings.

For a running system with streaming input, for efficient inference, 
we can cache image feature maps of the past frames.
For each moment, we only need to process images of current frame $t$ with backbone to get $\mathcal{F}_{\text{img}}^{(t)}$.
Feature maps of past frames $\{\mathcal{F}_{\text{img}}^{(t-1)}, \mathcal{F}_{\text{img}}^{(t-2)}, ...\}$ are directly fetched from cache, avoiding duplicated computation. 
Since most computation cost lies in backbone, \thename-T\ can run at a similar FPS compared with \thename.

\section{Experiments}
\subsection{Dataset}
We validate the effectiveness of \thename\ on the large-scale nuScenes~\cite{nuScenes} dataset, which is currently the most popular benchmark for vision-based methods.
NuScenes contains $1000$ driving sequences, with $700$, $150$ and $150$ sequences for training, validation and testing, respectively. Each sequence is approximately 20-second long and provides $6$ surround-view images per frame with the resolution of $1600\times900$. 
We submit \textit{test} set results to the online server for official evaluation to get the leaderboard results.
And other experiments are evaluated on the \textit{val} set.

\subsection{Experimental Settings}
We implement \thename\ with PyTorch~\cite{pytorch} framework and MMDetection3D~\cite{mmdet3d} toolbox. The results are based on three backbone configurations. We adopt ImageNet~\cite{imagenet} pretraining for ResNet-50,  FCOS3D~\cite{FCOS3D} pretraining for ResNet-101, and 
DD3D~\cite{DD3D} pretraining for VoVNet.
Unless specified, we adopt ResNet-50 as backbone for ablation experiments.
We train \thename\ on eight RTX3090 GPUs with the total batch size $8$. 
Inference speeds of all models are measured on one RTX3090 GPU.
We adopt mixed precision (float32 and float16) to accelerate training and disable it when measuring the inference speeds.
We do not use test-time augmentation or model ensemble during inference.
We adopt the implementation of \cite{DCNv2} for the context point sampling. The number of attention heads is set to $8$ as the common practice.
Results of \thename{}-T are based on only one past frame. Using more past frames are feasible and  brings more gain. More details about experimental settings will be available in the code.

In addition, we simply extend \thename\ for 3D object tracking by leveraging the tracking-by-detection algorithm~\cite{CenterPoint}.
Specifically, we project objects of the current frame back to the previous frame with the predicted velocity, and then match them with the tracked objects by closest distance matching. 
We evaluate the tracking performance on nuScenes tracking benchmark.

\subsection{Metrics}

\paragraph{\textbf{Detection.}} For 3D object detection, we report the official metrics:  NuScenes Detection Score (NDS), mean Average Precision (mAP) and true positive (TP) metrics (including Average Translation Error (ATE), Average Scale Error (ASE), Average Orientation Error (AOE), Average Velocity Error (AVE) and Average Attribute Error (AAE)). 
The main metric, NDS, is a weighted sum of the other metrics for comprehensively judging the detection capacity, defined as,
\begin{equation}
\begin{aligned}
\mathrm{NDS} = \frac{1}{10}\ [\ 5\mathrm{mAP}+\sum_{\mathrm{mTP}\in \mathbb{TP}}(1-\min(1, \mathrm{mTP}))\ ].
\end{aligned}
\end{equation}

\paragraph{\textbf{Tracking.}} For 3D object tracking, 
we report the official metrics: Average Multi Object Tracking Accuracy (AMOTA), Average Multi Object Tracking Precision (AMOTP), False Positives (FP), False Negatives (FN), Identity Switches (IDS), Track Initialization Duration (TID) and Longest Gap Duration (LGD).
AMOTA serves as the main metric, which penalizes ID switches, false positive, and false negatives and is averaged among various recall thresholds.

\begin{table}
    \small
    \centering
    % \renewcommand{\tabcolsep}{2.5pt}
% 	\begin{center}
   	\caption{\textbf{Performance comparison (\textit{val} set).} \thename\  achieves promising performance-speed trade-off. FPS are measured on one RTX3090 GPU.}	
   	\label{tab:res50_101}
    \resizebox{\linewidth}{!}{
	\begin{tabular}{l || c c || c c  c c c c c c}
        \toprule
		Method & Backbone & Input & \textbf{NDS}$\uparrow$ & FPS$\uparrow$ & mAP$\uparrow$ & mATE$\downarrow$ & mASE$\downarrow$ & mAOE$\downarrow$ & mAVE$\downarrow$ & mAAE$\downarrow$ \\
		\midrule
		DETR3D~\cite{DETR3D} & Res-50 & 1600 $\times$ 900 & 0.373 & 6.3 & 0.302 & 0.811 & 0.282 & 0.493 & 0.979 & 0.212\\
		BEVDet~\cite{BEVDet} & Res-50 & 704  $\times$ 256 & 0.372 & 7.8 & 0.286 & 0.724 & 0.278 & 0.590 & 0.873 & 0.247\\
		BEVDet~\cite{BEVDet} & Res-50 & 1056 $\times$ 384 & 0.381 & 4.2 & 0.304 & 0.719 & 0.272 & 0.555 & 0.903 & 0.257\\
	    \thename & Res-50 & 1600 $\times$ 900 & 0.409 & 6.0 &  0.338 & 0.768 & 0.284 & 0.443 & 0.883 & 0.221\\
        \thename-T & Res-50 & 1600 $\times$ 900 & 0.458 &  $\sim$6.0 & 0.354 & 0.748 & 0.277 & 0.432 & 0.539 & 0.197\\
		\midrule
		FCOS3D~\cite{FCOS3D} & Res-101 & 1600 $\times$ 900 & 0.372 & 1.7 & 0.295 & 0.806 & 0.268 & 0.511 & 1.315 & 0.170\\
        PGD~\cite{PGD} & Res-101 & 1600 $\times$ 900 & 0.409 & 1.4 & 0.335 & 0.732 & 0.263 & 0.423 & 1.285 & 0.172\\
		DETR3D~\cite{DETR3D} & Res-101 & 1600 $\times$ 900 & 0.425 & 3.7 & 0.346 & 0.773 & 0.268 & 0.383 & 0.842 & 0.216\\
	
		BEVDet~\cite{BEVDet} & Res-101 & 704  $\times$ 256 & 0.373 & 7.1 & 0.288 & 0.722 & 0.269 & 0.538 & 0.911 & 0.270\\
		BEVDet~\cite{BEVDet} & Res-101 & 1056 $\times$ 384 & 0.389 & 3.8 & 0.317 & 0.704 & 0.273 & 0.531 & 0.940 & 0.250\\
	    \thename & Res-101 & 1600 $\times$ 900 & 0.444 & 3.5 & 0.365 & 0.742 & 0.269 & 0.350 & 0.829 & 0.197\\
	    \thename-T & Res-101 & 1600 $\times$ 900 & 0.488 & $\sim$3.5 & 0.383 & 0.707 & 0.269 & 0.344 & 0.518 & 0.196\\
		\midrule
		DETR3D~\cite{DETR3D} & VoVNet & 1600 $\times$ 900 & 0.509 & 2.7 & 0.445 & 0.687 & 0.261 & 0.271 & 0.727 & 0.191\\
		\thename & VoVNet & 1600 $\times$ 900 & 0.532 & 2.5 & 0.462 & 0.628 & 0.262 & 0.263 & 0.658 & 0.180\\
		% & 0.444 & 3.5 & 0.365 & 0.742 & 0.269 & 0.350 & 0.829 & 0.197\\
		\bottomrule
	\end{tabular}
	}
\end{table}

\subsection{Main Results}
\paragraph{\textbf{Performance Comparison.}} 
In Tab.~\ref{tab:res50_101}, we compare \thename\ with other state-of-the-art methods on three backbone configurations. On both ResNet-50 and ResNet-101, \thename\ significantly outperforms DETR3D~\cite{DETR3D} and BEVDet~\cite{BEVDet} while achieving comparable inference speed.
And on ResNet-101, \thename\ outperforms FCOS3D~\cite{FCOS3D} and PGD~\cite{PGD} in terms of both performance and speed. 
On VoVNet, \thename\ significantly outperforms DETR3D~\cite{DETR3D}.
With temporal information, \thename{}-T achieves much higher results than \thename{}, especially in terms of mAVE. 

\paragraph{\textbf{NuScenes  Leaderboard.}}
Tab.~\ref{tab:det_leaderboard} shows the  3D detection leaderboard of nuScenes benchmark~\cite{nuScenes} for camera modality at the submission time (Mar. 4th,\ 2022). For fair comparison, we adopt the pretrained VoVNet~\cite{VOVNet} (refer to DD3D~\cite{DD3D}) as backbone,
the same with the other top methods on the leaderboard.
\thename\ \textbf{ranks 1st} on this highly competitive leaderboard.
And it's worth noting that 
our implementation is compact and elegant,
without test-time augmentation, multi-model ensemble or other tricks.

Tab.~\ref{tab:track_leaderboard} shows the tracking leaderboard of nuScenes benchmark~\cite{nuScenes} for camera modality (Mar. 4th,\ 2022). At the submission time, \thename\  \textbf{ranks 1st} on the leaderboard and  outperforms other methods by a large margin.
We only adopt a simple algorithm to generate tracking results. The tracking performance could be further improved with well-designed tracking algorithm.

\begin{table}
    \small
    \centering
    \renewcommand{\tabcolsep}{2.5pt}
% 	\begin{center}
   	\caption{\textbf{3D detection leaderboard of nuScenes benchmark (\textit{test} set) (Mar. 4th,\ 2022).} \thename\ \textbf{ranks 1st} on the highly competitive leaderboard at the submission time.}	\label{tab:det_leaderboard}
    % \resizebox{\linewidth}{!}
	\begin{tabular}{l c c c c c c c}
        \toprule
		Method & \textbf{NDS}$\uparrow$  & mAP$\uparrow$ & mATE$\downarrow$ & mASE$\downarrow$ & mAOE$\downarrow$ & mAVE$\downarrow$ & mAAE$\downarrow$ \\
		\midrule
		DD3D & 0.477 & 0.418 & 0.572 & 0.249 & 0.368 & 1.014 & 0.124\\
		DETR3D & 0.479 & 0.412 & 0.641 & 0.255 & 0.394 & 0.845 & 0.133\\
		BEVDet & 0.488 & 0.424 & 0.524 & 0.242 & 0.373 & 0.950 & 0.148\\
		TPD-e & 0.488 & 0.440 & 0.534 & 0.248 & 0.391 & 0.998 & 0.146\\
		\thename & 0.493 & 0.431 & 0.588 & 0.253 & 0.408 & 0.845 & 0.129\\
		\bottomrule
	\end{tabular}
\end{table}

\begin{table}
    \small
    \centering
    \renewcommand{\tabcolsep}{3.5pt}
% 	\begin{center}
   	\caption{\textbf{3D tracking leaderboard of nuScenes benchmark  (\textit{test} set)  (Mar. 4th,\ 2022).}}	\label{tab:track_leaderboard}
    % \resizebox{\linewidth}{!}
	\begin{tabular}{l c c c c c c c}
        \toprule
		Method & \textbf{AMOTA}$\uparrow$ &  AMOTP$\downarrow$ & FP$\downarrow$ & FN$\downarrow$ & IDS$\downarrow$ & TID$\downarrow$ & LGD$\downarrow$\\
		\midrule
		DEFT & 0.177 & 1.564 & 22163 & 60565 & 6901 & 1.600 & 3.080\\
		QD-3DT & 0.217 & 1.550 & 16495 & 60156 & 6856 & 1.620 & 2.961\\
		\thename & 0.273 & 1.185 & 18853 & 59150 & 2170 & 0.990 & 2.330\\
		\bottomrule
	\end{tabular}
\end{table}

\subsection{Ablation Study}

\paragraph{\textbf{Key Components.}}
We provide ablation experiments to validate the key components of \thename.
As shown in Tab.~\ref{tab:abla_key}, \POPs, context point and pixel ray respectively improve NDS by $1.5\%$, $0.9\%$ and $0.6\%$ and bring negligible overhead. It's worth noting that \POPs\ only reformulates the optimization problem without introducing any computational budget.
The significant improvement brought by \POPs\ validates its effectiveness. 
\begin{table}
    \small
    \centering
    \renewcommand{\tabcolsep}{2.5pt}
% 	\begin{center}
   	\caption{\textbf{Ablations about the key components of \thename.}}
	\label{tab:abla_key}
	\begin{tabular}{c c c c c c}
        \toprule
		- & Polar Param. &  Context Point & Pixel Ray & \textbf{NDS}$\uparrow$ & FPS$\uparrow$\\
		\midrule
	    A)  &  &  & & 0.373  & 6.3\\
	    B)  & \checkmark &  &  & 0.388  & 6.3\\
	    C)  & \checkmark & \checkmark &  & 0.397  & 6.0\\
	    D)  & \checkmark & \checkmark & \checkmark & 0.403  & 6.0\\
		\bottomrule
	\end{tabular}
    
\end{table} 

\paragraph{\textbf{Polar Velocity Decomposition.}}
Tab.~\ref{tab:velocity} presents the ablation study about the velocity decomposition.
We compare cartesian and polar decomposition based on  temporal input (frame $t-1$ and $t$). 
Polar decomposition results in much lower mAVE (mean Average Velocity Error) and achieves much accurate velocity estimation.
\begin{table}
    \small
    \centering
    \renewcommand{\tabcolsep}{2.5pt}
   	\caption{\textbf{Ablations about polar velocity decomposition.}}
	\label{tab:velocity}
    % \resizebox{\linewidth}{!}
	\begin{tabular}{c c c c}
    \toprule
    Velocity & mAVE $\downarrow$\\
    \midrule 
    Cartesian $(v_x, v_y)$ & 0.556\\
    Polar $(v_{rad}, v_{tan})$ & 0.539\\
	\bottomrule 
	\end{tabular}
\end{table}

\paragraph{\textbf{Scaling Factor.}}
Tab.~\ref{tab:scaling} shows the ablations about the scaling factor $k_{\text{scaling}}$.
Without numerically scaling up azimuth, \ie, $k_{\text{scaling}}=1$, the performance is poor because of the numerical unbalance between the tangential and radial direction.
We  use coarse grid search to tune $k_{\text{scaling}}$, and find $k_{\text{scaling}}=20$ corresponds to relatively good results. With finer tuning, higher performance can be expected.

\begin{table}
    \small
    \centering
    \renewcommand{\tabcolsep}{2.5pt}
% 	\begin{center}
   	\caption{\textbf{Ablations about the scaling factor.}}
	\label{tab:scaling}
    % \resizebox{\linewidth}{!}
	\begin{tabular}{c c c c c c c c}
        \toprule
		$k_{\text{scaling}}$ & \textbf{NDS}$\uparrow$ & mAP$\uparrow$ & mATE$\downarrow$ & mASE$\downarrow$ & mAOE$\downarrow$ & mAVE$\downarrow$ & mAAE$\downarrow$ \\
		\midrule
		1 & 0.294 & 0.235 & 0.935 & 0.292 & 0.718 & 1.073 & 0.290\\
		10 & 0.389 & 0.319 & 0.806 & 0.285 & 0.476 & 0.906 & 0.235\\
		20 & 0.403 & 0.330 & 0.771 & 0.277 & 0.459 & 0.873 & 0.237\\
		30 & 0.398 & 0.322 & 0.789 & 0.282 & 0.449 & 0.882 & 0.226\\
		\bottomrule
	\end{tabular}
\end{table} 

\paragraph{\textbf{Context Point.}}
Tab.~\ref{tab:point_num} shows the ablations about the number of context points.
Without context points, NDS is relatively low ($0.390)$ because of lack of contextual information.  Performance improves with the number increasing. But the gain get saturated with $4$ context points ($\text{NDS} = 0.403$). And further increasing the number has negative effects. By default $4$ context points are adopted in \thename.

In Tab.~\ref{tab:point_input}, we further explore the impact of input for generating context points.
As mentioned in Fig.~\ref{fig:Framework} and Eq.~(\ref{eq:offset}), context points are predicted based on a combination of query embeddings $q_i$ and center features $\{f_{c_i^1},...,f_{c_i^K}\}$. The results prove that both terms contribute to better context feature aggregation and higher performance.

\begin{table}
    \small
    \centering
    \renewcommand{\tabcolsep}{2.5pt}
% 	\begin{center}
   	\caption{\textbf{Ablations about the number of context points.}}
	\label{tab:point_num}
    % \resizebox{\linewidth}{!}
	\begin{tabular}{c c c c c c c c}
        \toprule
		Point & \textbf{NDS}$\uparrow$ & mAP$\uparrow$ & mATE$\downarrow$ & mASE$\downarrow$ & mAOE$\downarrow$ & mAVE$\downarrow$ & mAAE$\downarrow$ \\
		\midrule
		0 & 0.390 & 0.324 & 0.797 & 0.279 & 0.546 & 0.883 & 0.215\\
		2 & 0.402 & 0.329 & 0.787 & 0.281 & 0.466 & 0.869 & 0.223\\
        4 & 0.403 & 0.330 & 0.771 & 0.277 & 0.459 & 0.873 & 0.237\\
	    8 & 0.399 & 0.326 & 0.787 & 0.279 & 0.460 & 0.880 & 0.236\\
		\bottomrule
	\end{tabular}
	\vspace{-20pt}
\end{table} 

\begin{table}
    \small
    \centering
    \renewcommand{\tabcolsep}{2.5pt}
% 	\begin{center}
   	\caption{\textbf{Ablations about the input for generating context points.}}
	\label{tab:point_input}
    % \resizebox{\linewidth}{!}
	\begin{tabular}{c c c c}
        \toprule
	    Query Embeddings & Center Features  & \textbf{NDS}$\uparrow$ & mAP$\uparrow$ \\ 
	    \midrule
	    \checkmark & & 0.393 & 0.324 \\
	    & \checkmark & 0.403 & 0.325 \\
	    \checkmark & \checkmark & 0.403 & 0.330 \\
		\bottomrule 
	\end{tabular}
\end{table}

\paragraph{\textbf{Decoder Layer.}}
Ablations about the number of decoder layers  are shown  in Tab.~\ref{tab:layer}. With only  $1$ decoder layer (no iteration), results are poor. 
In \thename, we adopt $6$ decoder layers, which corresponds to relatively good results.

\begin{table}[t]
    \small
    \centering
    \renewcommand{\tabcolsep}{2.5pt}
% 	\begin{center}
    % \vspace{-20pt}
   	\caption{\textbf{Ablations about the number of decoder layers.}}
	\label{tab:layer}
    % \resizebox{\linewidth}{!}
	\begin{tabular}{c c c c c c c c}
        \toprule
		Layer & \textbf{NDS}$\uparrow$ & mAP$\uparrow$ & mATE$\downarrow$ & mASE$\downarrow$ & mAOE$\downarrow$ & mAVE$\downarrow$ & mAAE$\downarrow$ \\
		\midrule
		1 & 0.256 & 0.233 & 0.901 & 0.318 & 1.167 & 1.182 & 0.389\\
		2 & 0.358 & 0.316 & 0.792 & 0.286 & 0.662 & 1.044 & 0.262\\
		3 & 0.383 & 0.328 & 0.790 & 0.286 & 0.532 & 0.957 & 0.243\\
	    4 & 0.395 & 0.323 & 0.786 & 0.277 & 0.461 & 0.890 & 0.248\\
	    5 & 0.401 & 0.329 & 0.769 & 0.277 & 0.472 & 0.884 & 0.240\\
	    6 & 0.403 & 0.330 & 0.771 & 0.277 & 0.459 & 0.873 & 0.237\\
	    7 & 0.398 & 0.325 & 0.780 & 0.280 & 0.440 & 0.904 & 0.235\\
		\bottomrule
	\end{tabular}
\end{table}

\begin{figure}[]
    \centering
    \includegraphics[width=\linewidth]{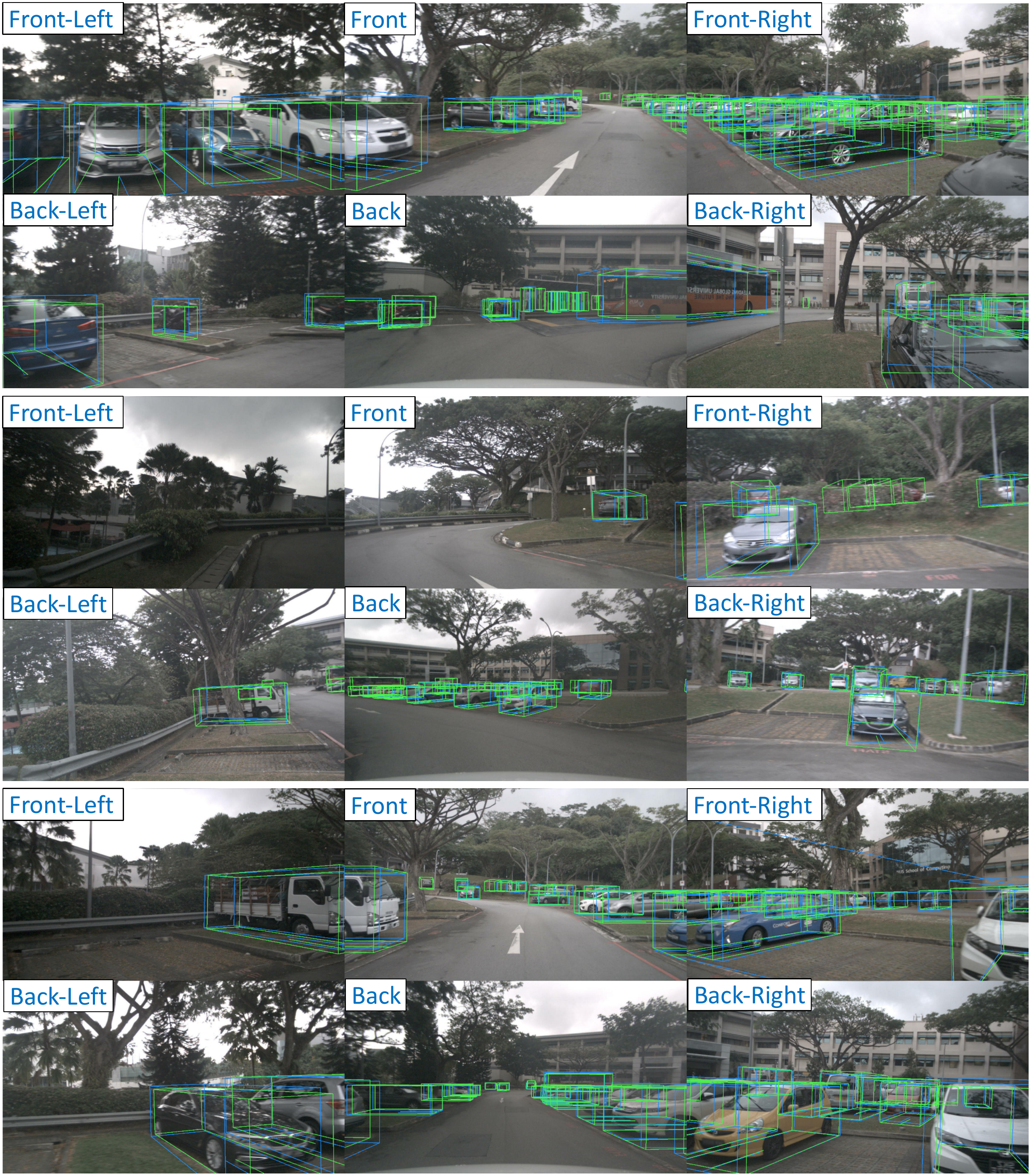}
    \caption{\textbf{Qualitative results about final predictions.} For visualization, 3D bounding box predictions are projected onto surround-view images. Blue ones denote predictions while green ones denote GTs.}
    \label{fig:prediction}
\end{figure}

\begin{figure}[]
    \centering
    \includegraphics[width=\linewidth]{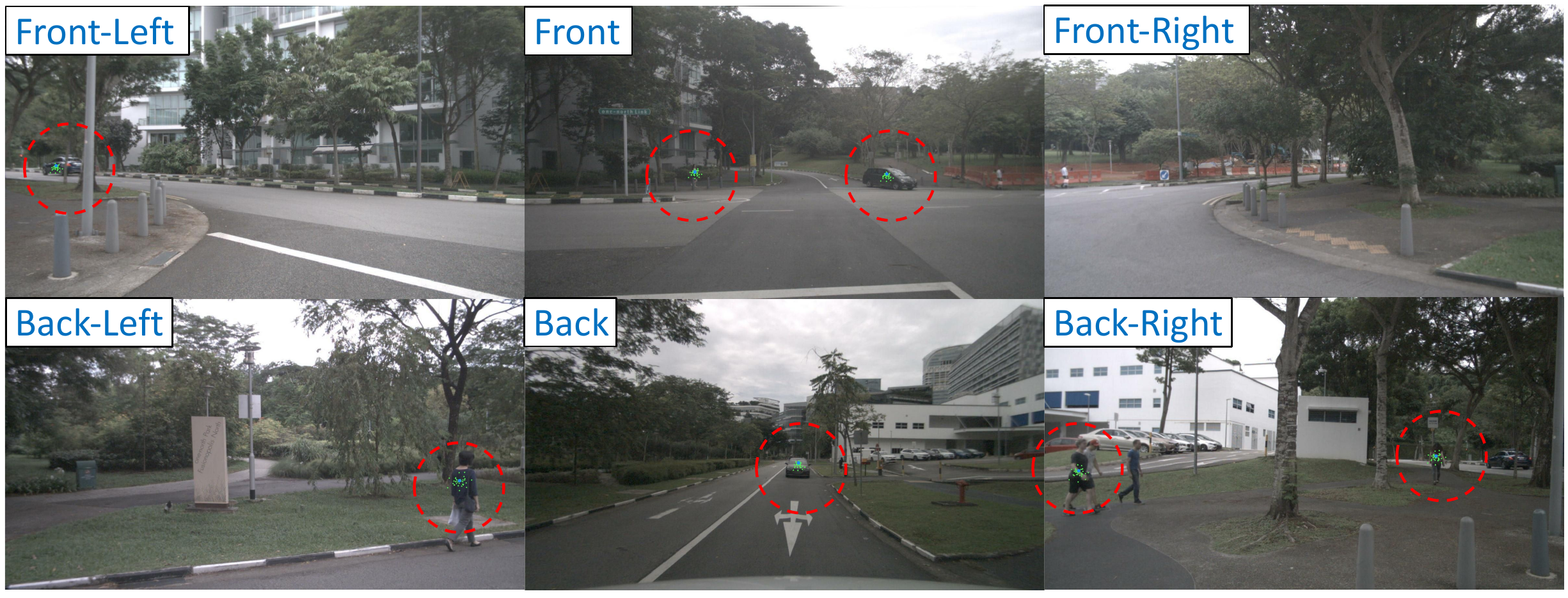}
    \caption{\textbf{Visualizations about center (blue) and context (green) points.} Several queries are selected for presentation (circled with red). Please zoom in for better view.}
    \label{fig:aggregation}
\end{figure}

\section{Qualitative Results}
\paragraph{\textbf{Center-Context Feature Aggregation.}}
Visualizations about center and context points are in Fig.~\ref{fig:aggregation}.
Center points focus on object's center for localization while context points focus on a wider region to capture more information. They complement each other and contribute to better feature aggregation.

\paragraph{\textbf{Detection Results.}}
Qualitative results about final predictions are shown in Fig.~\ref{fig:prediction}. Blue and green boxes respectively denote predictions and GTs. 
Highly occluded or far-away objects may result in bad cases, which is a common problem faced by all detectors. 
Expect these challenging cases,  \thename\ achieves stable and satisfactory detection results.
 
\section{Conclusion}
In this paper, we present \POPs{} to exploit the view symmetry of surround-view camera system.  \POPs{} establishes explicit associations between image patterns and
prediction targets, superior to \IMPs{} and \CAPs{}.
Based on \POPs{}, \thename\ achieves promising performance in terms of both 3D  detection and 3D tracking on the challenging nuScenes benchmark.
And \POPs{} can be extended to other perception task and even planning task. We leave it as further work.
\clearpage
\bibliographystyle{splncs04}
\bibliography{main}
\end{document}